\documentclass[a4paper,conference]{IEEEtran}
\usepackage[font=small,labelfont=bf]{caption}
\usepackage{microtype}
\usepackage{xspace}
\usepackage{epsfig}
\usepackage{graphicx}
\usepackage{amsmath, bm}
\usepackage{amssymb}
\usepackage{algorithm}
\usepackage{algorithmic}
\usepackage{mathtools}
\usepackage{array}
\usepackage{multirow}
\usepackage[inline]{enumitem}
\usepackage{multirow}
\usepackage{tabularx}
\usepackage{color}
\usepackage[utf8]{inputenc} % allow utf-8 input
\usepackage[T1]{fontenc}    % use 8-bit T1 fonts
\usepackage{booktabs}       % professional-quality tables
\usepackage{amsfonts}       % blackboard math symbols
\usepackage{nicefrac}       % compact symbols for 1/2, etc.
\usepackage{microtype}      % microtypography
\usepackage{rotating}
\usepackage{lipsum}
\usepackage{wrapfig}
\usepackage{tikz}
\usepackage{flushend}
\usepackage{cool}
\usepackage{siunitx}
\usepackage{bbm}

\usepackage[capitalise,nameinlink]{cleveref}

\makeatletter
\crefname{section}{\S\@gobble}{\S\S\@gobble}
\Crefname{section}{\S\@gobble}{\S\S\@gobble}
\makeatother
\makeatletter
\DeclareRobustCommand\onedot{\futurelet\@let@token\@onedot}
\def\@onedot{\ifx\@let@token.\else.\null\fi\xspace}
\makeatother

\def\etc{\emph{etc}\onedot}
\def\ie{\emph{i.e}\onedot}
\def\eg{\emph{e.g}\onedot}
\def\cf{\emph{cf}\onedot}
\def\vs{\emph{vs}\onedot}

\def\R{\mathbb{R}}

\definecolor{redcol}{rgb}{1, 0, 0}
\definecolor{bluecol}{rgb}{0, 0, 1}

\renewcommand{\paragraph}[1]{\smallskip\noindent{\bf{#1}}}
\newcommand{\approach}[0]{GraphMapper\xspace}
\newcommand{\spl}{\texttt{SPL}\xspace}
\newcommand{\sr}{\texttt{SR}\xspace}

\newcommand{\rgb}{\texttt{RGB}\xspace}
\newcommand{\depth}{\texttt{Depth}\xspace}

\DeclareSIUnit{\million}{\text{milion}}

\usepackage{pgfplots}
\usepackage{wrapfig}

\usepackage{xr}
\usepackage{footnote}
\usepackage[skip=2pt,font=small]{caption}

\pgfplotsset{compat=1.16}

\ifCLASSINFOpdf
\else
\fi

\title{\approach: Efficient Visual Navigation by Scene Graph Generation}

\author{\IEEEauthorblockN{Zachary Seymour, Niluthpol Chowdhury Mithun, Han-Pang Chiu, Supun Samarasekera, Rakesh Kumar}

\IEEEauthorblockA{Center for Vision Technologies, SRI International, Princeton, NJ; \ Email: firstname.lastname@sri.com}
}

\begin{document}

\maketitle

%{
% \renewcommand{\thefootnote}%
%    {\fnsymbol{footnote}}
% \footnotetext[2]{\scriptsize Center for Vision Tech., SRI International; {\tt\scriptsize firstname.lastname@sri.com}}
%}
\begin{abstract}
Understanding the geometric relationships between objects in a scene is a core capability in enabling both humans and autonomous agents to navigate in new environments. A sparse, unified representation of the scene topology will allow agents to act efficiently to move through their environment, communicate the environment state with others, and utilize the representation for diverse downstream tasks. To this end, we propose a method to train an autonomous agent to learn to accumulate a 3D scene graph representation of its environment by simultaneously learning to navigate through said environment. We demonstrate that our approach, GraphMapper, enables the learning of effective navigation policies through fewer interactions with the environment than vision-based systems alone. Further, we show that GraphMapper can act as a modular scene encoder to operate alongside existing Learning-based solutions to not only increase navigational efficiency but also generate intermediate scene representations that are useful for other future tasks.
\end{abstract}

%-------------------------------------------------------------------------
\section{Introduction}

Automatic understanding of geometric relationships among semantic objects in the perceived scene is a key ability for humans to navigate in new environments. Humans reason about the 3D topology of recognized objects in the space to enable efficient navigation, such as finding the shortest path from one room to another avoiding obstacles. Human brains automatically builds a unified representation of the environment to support memory and guide future actions for navigation tasks.   

\iffalse
\begin{figure}[t]
    \centering
    \includegraphics[width=0.6\columnwidth]{figures/sgtn.png}
    \vspace{0.1cm}
    \caption{A high-level overview of our proposed architecture, \approach. Given bounding boxes of scene objects and estimates of their 3D positions, our Scene Graph Transformer Network emits a graph representation of the scene, along with node- and edge-level features, to enable visual navigation.}
    \label{fig:graphmapper}
\vspace{-0.5cm}
\end{figure}
\fi

The goal for this paper is to train an autonomous agent to resemble this human capability for efficient visual navigation. Training agents with intermediate scene representations ought to enable the agents to learn faster and are more robust under different situations~\cite{representation}. Recently, deep reinforcement learning (DRL) methods demonstrated promising results for autonomous agents on navigation tasks, by learning a direct mapping from observations to actions through trial-and-error interactions with its environments. However, most of these methods are purely data driven without constructing intermediate representations. Generating a powerful but cost-effective scene representation, that encodes both semantic objects and their geometric relationships, is still a challenging and unsolved problem.
%in this field.

\begin{figure}[t]
    \centering
    \includegraphics[width=0.9\columnwidth]{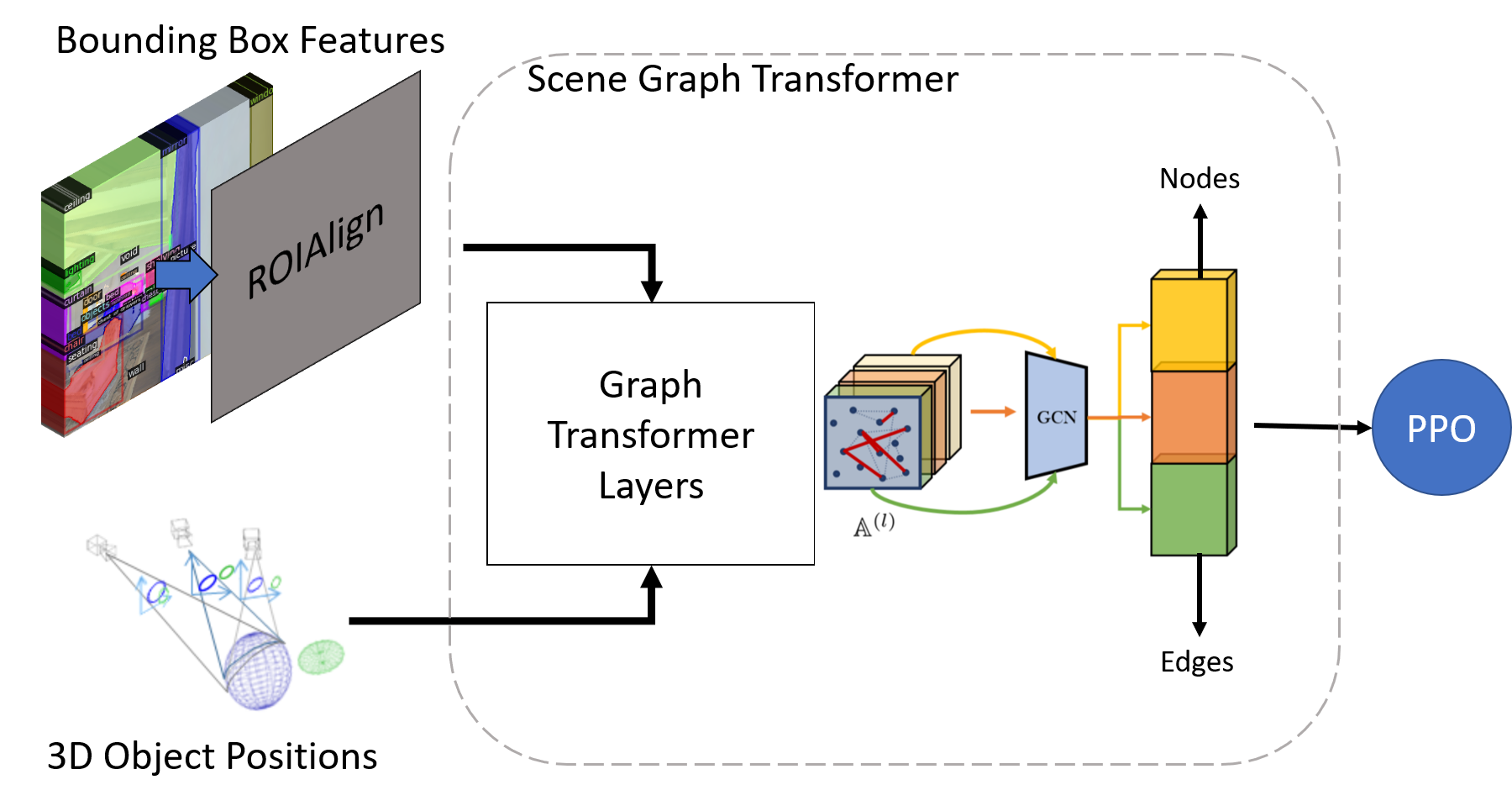}
    \caption{An overview of proposed \approach. Given bounding boxes of scene objects and estimates of their 3D positions, our Scene Graph Transformer Network emits a graph representation of the scene, along with node- and edge-level features, to enable visual navigation.}
    \vspace{-0.3cm}
    \label{fig:graphmapper}
\end{figure}

We propose to learn and to generate “scene graphs” for fulfilling this purpose. 
Scene graphs are compact representations of the relationships between the objects in a scene or image, using a graph data structure.
Each node in the graph represents an instance of an object in the scene, and each edge encodes the relationship between a pair of objects.
Scene graphs offer an appealing mid- to high-level representation of the scene that are easily digestible by many applications, and graph-structured data in general has been shown to act as a strong prior to regularize and refine features in neural networks~\cite{battaglia18relinducbias}.
We adapt the recently introduced Graph Transformer Network (GTN)~\cite{je19GTN} to the task of simultaneous scene graph generation and visual information refinement for building up powerful scene representations to perform visual navigation, which we refer to as \approach (\Cref{fig:graphmapper}).
Beginning with a candidate set of scene objects and simple, pairwise edges defining the geometric relationship of each, we learn sparse, structural scene graph representations that not only improve the sample efficiency of the learning policy but also afford many opportunities for down-stream tasks and future applications. The contributions of proposed \approach, are as follows:
\begin{enumerate}
    \item We are the first such work to explore the generation of 3D scene graphs as an additional output alongside  training an autonomous agent to perform visual navigation.
    \item We offer a widely-applicable and easily-adaptable "renderer-supervised" approach that alleviates the need for large-scale, annotation of scene graph ground truth.
    \item We show how the graph-structured data refines the agent's visual observations, leading to greater performance over the same training regime on practical learning tasks, i.e., PointGoal and Vision Language navigation (VLN).
    \item Finally, we explore several representative examples of the scene representations the agent learns to predict that hint at several interesting uses in future applications.
\end{enumerate}

\section{Related Work}
%Autonomous navigation in large, unknown environments is an important and challenging problem to many applications. Here
We briefly review prior works related to different learning-based approaches to the problem of autonomous navigation, with the discussion on scene representations in these methods.

%\noindent \textbf{Traditional Approaches:} The traditional (non-learning) approach decomposes this problem into two steps. The first step is mapping ~\cite{thrun2005probabilistic, davison1998mobile, desouza2002vision}: a highly accurate geometry map is constructed using simultaneous localization and mapping (SLAM) algorithms during navigation. The second step is path planning ~\cite{lavalle2000rapidly,canny1988complexity,kavraki1996probabilistic}: paths are planned through the constructed map afterwards. However, this approach requires hand-engineering and is not always feasible in new unexplored environments. 

\noindent \textbf{Learning-based Approaches:} The traditional (non-learning) approach decomposes this problem into two steps, i.e., mapping ~\cite{thrun2005probabilistic, davison1998mobile, desouza2002vision,krasner2022signav} and path planning~\cite{lavalle2000rapidly,canny1988complexity,kavraki1996probabilistic}. Instead of dividing the problem into two discrete stages, learning based methods~\cite{jaderberg2016reinforcement, pathak2017curiosity,gupta2017unifying,mnih2016asynchronous,exp4nav,AI2THOR,dosovitskiy2016learning,anm20iclr} address it by learning direct correspondences from sensor data to robot motion commands as navigation policies. Due to the difficulty of training agents through interactions with real world, most of the methods focus on training using photo-realistic simulators ~\cite{habitat19iccv,Mishkin2019BenchmarkingCA,MidLevelVisualRep,situationalfusion}. 
However, this requires significant training data (experience) to implicitly learn the knowledge to achieve better navigation results ~\cite{Mishkin2019BenchmarkingCA,habitat19iccv,ddppo} compared to traditional methods in new environments.
Moreover, it lacks capabilities to explain or reason its behaviors or actions.
While recent works have shown promise in leveraging a hierarchical policy combined with an analytical planner~\cite{anm20iclr,occant,chaplot20objectgoal}, our focus remains on improving the efficiency of end-to-end learning based solutions by improving intermediate representations.

\noindent \textbf{Scene Representations:} There are recent works ~\cite{exp4nav,savinov2018episodic,gupta17:cogmapping,anm20iclr,semmapnet} that aim to combine the strengths from traditional SLAM-based navigation methods and data-driven  learning-based methods. These works build up some forms of top-down occupancy map of the environment, which the agent either uses as an additional feature for its policy, or as an environmental proxy to select intermediate goals, or both. By utilizing this occupancy map, the agent is aware of its options ~\cite{yamauchi1997frontier,thrun1999minerva} to explore more areas for navigation tasks. Building this kind of representation helps the agent to learn more effectively, by increasing the sample efficiency of learning the policy and reducing the amount of training data for visual navigation~\cite{anm20iclr,exp4nav,MidLevelVisualRep,maast}. While utilizing such a representation in deep reinforcement learning (DRL) or imitation learning (IL) based methods has similar benefits that a geometric map typically supports in traditional SLAM, it still lacks the cognitive capability that the humans possess to understand the 3D topology among semantic objects in the environment. As such, we seek to develop a richer, sparser scene representation specifically for autonomous agents in navigation tasks. Our representation is in the form of a scene graph that encodes entities, semantic objects, and their relationships in the environment.
Of note, we distinguish our work from those in visual navigation which utilize or construct environment-level topological graphs~\cite{savinov2018semi, beeching2020learning, wang2021structured, chaplot2020neural} in that our focus is on generating a scene-level graph from a particular viewpoint.

\noindent \textbf{Scene Graph Generation and Graph Neural Networks:} Scene graph representations serve as a powerful way of representing image content in a heterogeneous graph, in which the nodes represent objects or regions in an image and the edges represent relationships between the objects.
Scene graphs have largely been studied as a higher-level, more descriptive form of image annotation (see~\cite{li17scenegraphgeneration}); however, they have also been recently studied in other applications, such as image synthesis~\cite{johnson18imagesyntheis, mo2019structurenet, ashual2019specifying, qi2018human}.
Similarly, graph neural networks (GNN)---of which graph convolutional networks (GCN)~\cite{GCN} seem to be the most popular type---are a new class of neural networks that allow the processing of graph-structured data in a deep learning framework.
Such networks work by learning a representation for the nodes and edges in the graph and iteratively refining the representation via ``message passing;'' \ie, sharing representations between neighboring nodes in the graph, conditioned on the graph structure.

The majority of works on scene graph generation focus on 2D scenes: given a set of objects in an image, the goal is to either predict a relationship between object pairs (\cf,~\cite{desai10graph,lu16vrd}) or to transform a complete graph of the scene into a sparser set of semantic relationships (\cf, \cite{sgg_imp,graph_rcnn,factorizablenet}).
A few recent works instead pose the scene graph generation problem in a less anthropocentric way: they focus on finding graphs describing the geometric structure of the scene, often in 3D~\cite{scenegraphnet,3dscenegraph,vgfm}. 
However, such methods generally require learning a separate function for each type of edge in order to perform message passing in a manner that treats different types of edge relationships differently.
The computation becomes expensive when many nodes or edge types exist, and the GNN computation can become too heavy to act as a component in a larger system.
To enable end-to-end graph generation and graph feature learning, we leverage Graph Transformer Networks~\cite{je19GTN} to build up complex edges, by forming ``meta-paths'' from given an initial set simple geometric edge relationships. 
This network learns to classify nodes and edges, while performing feature message passing using lightweight GCN operations.
%\section{Approach}
\section{Problem Setup}
To verify and demonstrate GraphMapper, we train our proposed architecture for PointGoal and VLN tasks. We assume our agent is equipped only with \rgb camera and a \depth sensor, in addition to some means of obtaining region of interest (ROI) proposals for its current view.
%We assume the agent has a sensor capable of producing a set of ROIs for a given \rgb observation. 
Here we follow ~\cite{vgfm} that utilizes semantic masks provided by the simulator's rendering engine as ROIs. In practice, these ROIs can be predicted on the fly by, \eg, an object detector~\cite{he20maskrcnn}. 

Our goal is to learn an action policy $\pi$ for the agent to utilize \rgb and \depth sensor data in a more effective manner, that enhances scene understanding and enables efficient navigation. 
To achieve this, we propose a new method that enables the agent to generate single-shot 3D scene graphs of the environment using a novel formulation of the recently-introduced Graph Transformer Networks~\cite{je19GTN}. 
In other words, by navigating through or exploring the environment, the agent learns to generate a graph representation from its current observation. 
The graph representation encodes the semantic objects in the scene and the 3D structural relationships among them. 
Prior work has shown that providing geometric information of the environment (such as a top-down occupancy map) helps the autonomous agent to learn more efficiently, to transfer more easily across domains, and to explore new environments more aggressively~\cite{exp4nav, anm20iclr}. 
Our scene graph representation models objects in the scene along with their geometric relationships. 
It fuses both the semantic entities and their geometry information into a single, computationally-efficient representation. 
This representation enables higher-level scene understanding that cannot be easily achieved with a top-down geometric map.

%Finally, we propose 
% These networks extract a global graph representation to represent the agent's current observation. They also enable efficient processing and prediction of multiple relationship types (graph edges) among objects, without the need for extensive human annotations.

\section{Renderer-Supervised Scene Graph Generation}
\label{sec:sgg}
There has been several recent works addressing scene graph generation; \ie, given one or more views of a scene, output a graph $G = (V,E)$ where $V$, the set of nodes, represents objects present in the scene and $E$, the set of edges, the relationships between them.
One issue in adapting these methods for visual navigation tasks, is that a number of these methods \begin{enumerate*}[label=\arabic*)]
    \item rely on a large quantity of manually-annotated, graph level supervision to train~\cite{graph_rcnn, sgg_imp, vgfm},
    \item output a large set of semantic relationship edges, many of which are meaningless in the context of autonomous agents (\eg, ``belonging to,'' ``wearing,'' ``playing'')~\cite{factorizablenet, graph_rcnn}, or
    \item have only been shown to work either in synthetic environments or in narrow fields-of-view unsuitable for visual navigation~\cite{vgfm, scenegraphnet}.
\end{enumerate*}

By contrast, we seek a solution for the generation of scene graphs that does not require any additional annotation, outputs only structurally-meaningful relationships, and operates on the fly from the point-of-view of an autonomous agent.
Our core intuition is that in the typical training environment for such an autonomous agent---\eg, Habitat~\cite{habitat19iccv}, Gibson~\cite{Gibsonenv}, MINOS~\cite{MINOS}, or AI2-THOR~\cite{AI2THOR}---there is a rich quantity of underlying data that goes into rendering the agent's current view.
Namely, given a view frustum derived from the agent's current pose (3D position and 3D orientation or angular heading) and camera extrinsics, the agent's \rgb observation is a slice of the underlying 3D mesh, which readily encodes structural information about the entire scene, often augmented with discrete object meshes.
%As such, we propose to make use of this information to conduct ``renderer-supervised'' scene graph generation.
We propose to use this information for ``renderer-supervised'' scene graph generation.

\vspace{0.05cm}
\noindent \textbf{Dataset Extraction.} First, we introduce our method for automatically extracting scene graph ground truth to supervise our scene graph generation model.
To do so, we make use of the underlying semantic annotations that are readily available in most of the popular learning environments~\cite{Gibsonenv,habitat19iccv}.
Such annotations, propagated from semantic point cloud or existing 3D models, are cheaper to collect than full annotations for large-scale environments~\cite{3dscenegraph}.
The objects defined in this scene become the nodes in our graph.
To provide structurally-meaning edges to the graph, we utilize the information already encoded in the 3D mesh.
Inspired by prior work on 3D scene graphs~\cite{vgfm, scenegraphnet, 3dscenegraph} and the types of edges used in other domains (\eg, Visual Genome~\cite{visualgenome}), we utilize two main types of edges: \emph{co-planarity}, with a three types of edge denoting two objects are approximately co-planar in the $x$, $y$, and $z$ directions; and \emph{same region}, where an edge represents that two objects share a region in the semantic scene (typically indicating ``same room,'' depending on the discretization of the space).
As we utilize the underlying 3D mesh and the environment's semantic scene, we can directly compute either oriented or axis-aligned bounding boxes (AABB) for objects in the environment, which makes computing this ground truth simple.
We find empirically that it adds relatively little time to the rendering of the environment, compared to loading semantic 3D mesh from disk.

\noindent \textbf{Graph-based Scene Observations.} Now, we discuss the input to our scene graph generation network that maps observations to the ground truth scene graphs.
As mentioned above, our agent's visual observation begins with a set of ROIs given its \rgb observation.
Each ROI corresponds to one of the objects defined in the given scene; namely, one of the defined Matterport object classes.
While earlier methods approach the problem of scene graph generation by predicting the existence of and label for an edge between each pair of objects independently, it has become common practice to construct a complete graph between all of the object nodes and utilize message passing to label the edges holistically.
That is, for a set of $N$ ROIs, a graph will be generated with $N(N-1)$ candidate edges.
As our graph relationships are essentially structural in nature rather than linguistic or semantic, we use the relative pose and dimensions of the observed objects as initial features for the candidate edges, similar to the quadric representation used in~\cite{vgfm}.
% To more realistically model the expected behavior of our approach outside of simulation, we make use of a realistic noisy depth model~\cite{Choi_2015_CVPR} and do not assume accurate localization of the agent.
We utilize the agent's depth observation to estimate both the dimensions of each object, as well as the approximate location of the object's front plane in the agent's coordinate frame.
We concatenate these two 6-dimensional vectors (3 dimensions and 3 coordinates) for each pair of nodes as the input edge representation.
As there is no assumed directionality to our edges (and thus no strict notion of ``subject'' or ``object'' in the predicted relationships as in prior work~\cite{graph_rcnn}), we adapt a newly-introduced GNN formulation to simultaneously extract meaningful edges from this fully-connected graph and to perform message passing across heterogeneous edge types.

\subsection{Scene Graph Transformer Network}
Most prior scene graph generation algorithms are built around the idea of iterative message passing~\cite{sgg_imp}, whereby a set of node features and candidate edge features are repeatedly propagated over the edges of the graph to eventually output the final relationship predictions.
However, because our goal is to learn to predict the graph edges while simultaneously utilizing the graph structure for performing the agent's downstream task, we cannot realistically afford such a complex operation.
Furthermore, because we leverage graph input extracted from the rendering engine rather than higher level annotations, the relationships represented by single-hop graph edges may be too simple for the node representations to contain meaningful global information.
Based on these two insights, we leverage Graph Transformer Networks (GTN)~\cite{je19GTN} as a model that can perform soft edge selection to represent composite relationships and handle the high-degree of heterogeneity present in our graph.
GTNs~\cite{je19GTN} were introduced as a graph variant of the Spatial Transformer Network~\cite{spatialtransformer} to automatically learn transformations of the graph adjacency matrix to encode ``meta-paths'' (paths comprised of heterogeneous edge types) between nodes. This transforms the input graph structure into one defined by these meta-paths of arbitrary length and edge type in a way that is more suitable for the downstream task. We provide a brief summary of the model below.
%provide a brief summary of the basic model in \Cref{app_sec:gtn} and 
%refer the reader to~\cite{je19GTN} for more architectural details.

%We provide a brief summary of the basic model and refer the reader to~\cite{je19GTN} for more details.

%\vspace{0.1cm}
\noindent \textbf{Notation.} The input to our Scene Graph GTN is a complete graph $G = (V, E)$ on the set of $V$ nodes.
If $N = |V|$ denote the number of nodes, then $E = \{(i, j) \mid \forall i,j \in V\}$ and $|E| = N(N-1)$.
Each $v \in V$ is represented by a vector $z \in \R^{d_V}$, where $d_V$ is the output dimension of CNN used to process the node ROIs.
Each $e \in E$ is represented by a vector $a \in \R^{d_E}$, where $d_E$ is the number of input edge features defined above; \ie, $d_E = 6$.
The structure of the input graph can then be compactly represented as tensor $A \in \R^{N \times N \times d_E}$. 
After several layers of processing, the output of this network is a set of $N$ labels, one for each node in the original graph, and a new, sparse adjacency matrix $A_s \in \{0, 1\}^{N \times N}$.
In experiments, we use three (3) layers for both the adjacency matrix in GTN and for node-node message passing in the following GCN.

%\vspace{0.1cm}
\noindent \textbf{Graph Transformer Layer.} The generation of new set of paths is performed by a sequence of Graph Transformer (GT) layers, each of which uses soft attention to select candidate graph structures initial complete adjacency matrix $A$. 
The layer outputs a new multi-hop adjacency matrix as the composition of the soft-selected candidate graph structures. 
It computes a transformation of adjacency matrix $A$ using a $1\times1$ convolution with the softmax of two sets of learned weights $W_1, W_2 \in \R^{1 \times 1 \times d_E}$ (where $W_1$ and $W_2$ parameterize the convolution operation) in a manner similar to Transformer's dot-product attention~\cite{vaswani17:attention}. 
A number of such layers can be stacked to generate arbitrary length $l$ paths with adjacency matrix $A_P$:
\begin{equation}
\footnotesize
{A}_{P} = \left(\sum_{t_1 \in \mathcal{T}^e}{\alpha_{t_1}^{(1)} A_{t_1}}\right) \left(\sum_{t_2 \in \mathcal{T}^e}{\alpha_{t_2}^{(2)} A_{t_2}}\right) \dotsm \left(\sum_{t_l \in \mathcal{T}^e}{\alpha_{t_l}^{(l)} A_{t_l}}\right)
\label{eq:gtn_adj}
\end{equation}
To ensure the network is able to learn meta-paths that are of any length and that may also include some of the original graph edges, the identity matrix is always included; \ie, $A_{0}=I$.

%\vspace{0.1cm}
\noindent \textbf{Graph Convolutional Network. }Following a sequence of Graph Transformer layers, a graph convolutional network (GCN)~\cite{GCN} is used to learn useful representations for node representations in an end-to-end fashion.
With $H^{(l)}$ the node features at the $l$th layer in the GCN, the message passing operation is performed as:
\begin{equation}
\footnotesize
{H}^{(l+1)} = \sigma \left (\Tilde{D}^{-\frac{1}{2}}\Tilde{A}\Tilde{D}^{-\frac{1}{2}}H^{(l)}W^{(l)} \right ), 
\label{eq:gcn}
\end{equation}
Here $\Tilde{A} = A+I \in \R^{N\times N}$ represents $A$ with added self-loops; $\tilde{D}$ is the degree matrix of $\tilde{A}$ (where $D_{ii}$ represents the degree of node $i$ and $W^{(l)} \in \R^{d \times d}$ is a learned parameter matrix for layer $l$). 
In standard GCN, only the node-wise linear transform $H^{(l)} W^{(l)}$ is learnable. 
However, the preceding GTN layers output several update (meta-path) adjacency matrices at each iteration, allowing the entire graph structure to be learned. 
If the number of output channels of $1\times1$ convolution in each GT layer is $C$, then each layer outputs a pair of intermediate adjacency tensors $Q_1$ and $Q_2 \in \R^{N \times N \times C}$.
This architecture then functions as as an ensemble of GCNs operating on each of $C$ adjacency matrices output by the final GT layer.
A GCN is applied to each channel $C$ and the multiple node representations are concatenated to form a holistic node representation $Z \in \R^{N \times C \cdot d}$, where $d$ is the node feature dimension.

For our purposes, we also preserve the output adjacency tensor $Q_1 \otimes Q_2=\hat{A} \in \R^{N \times N \times C}$ as an $C$-dimensional feature vector for each edge. 
We then utilize two separate fully-connected heads.
It first predicts a class for each node given $Z$, using standard cross-entropy given the objects present in the scene as supervision.
The other outputs zero or more types for each edge given $\hat{A}$, using multi-label binary cross-entropy given the edge types extracted from the renderer (as described in \Cref{sec:sgg}) as supervision.
The whole architecture also receives feedback from gradients propagated from the reward signal of the DRL or IL policy learning algorithm.

\section{Evaluation}
In this section, we demonstrate the effectiveness of our proposed \approach, for producing refined, structural visual representations for DRL-based PointGoal (\ref{sec:exp_drl}) and IL-based VLN navigation (\ref{sec:exp_il}) tasks. During navigation, the agent constructs and accumulates a scene graph representation for the perceived environment for more effective decision-making. Our scene graph representation also supports novel downstream applications, such as explaining or reasoning its behaviors or actions. 
Although we choose PointGoal and VLN as target tasks, our GraphMapper can be applied for any navigation task \cite{Anderson2018OnEO} that requires the agent to move throughout the environment, such as ObjectGoal, AreaGoal, or Exploration. Following prior works \cite{Anderson2018OnEO, mattersim, krantz2020navgraph}, we use Success weighted by Path Length (\spl) and Success Rate (\sr) as the primary metrics to evaluate the navigation performance of agents. We also report Normalized Dynamic Time Warping (NDTW) and Navigation Error (NE) for VLN agents~\cite{49206, krantz2020navgraph}.
%We first describe in detail the evaluation set-up, then introduce the baseline methods considered, and finally discuss the obtained empirical results as well as several qualitative applications.
Please see supplementary for more details on experiments, and qualitative applications.

\subsection{PointGoal Navigation } 
\label{sec:exp_drl}
\newcommand{\fcnn}{f_{\text{CNN}}}
\newcommand{\fmap}{f_{\text{MAP}}}
PointGoal navigation requires the agent to navigate to an assigned target position, given a starting location and orientation. 
%During navigation, the agent constructs and accumulates a scene graph representation for the perceived environment for more effective decision-making. 
%\paragraph{Policy Architecture, Rewards and Optimization. }
%An overall diagram of our policy architecture is shown in \Cref{fig:arch}.
Our base network for performing PointGoal navigation is derived from the ``RL + Res18'' baseline described in~\cite{anm20iclr}.
Namely, we replace the ResNet18~\cite{Resnet} visual encoder with our Scene Graph Transformer Network.
The features of each node are average pooled and passed through fully-connected layer, and then input along with the agent's previous action and an encoding of the target to a one-layer GRU~\cite{cho14:gru} to enable the agent to maintain coherent action sequences across several time steps (the actor), followed by a linear layer (the critic) to predict the next action. To increase sample efficiency, we follow~\cite{anm20iclr} and replace the continuous representation of the agent's relative angle and distance to the target and the one-hot encoding of its previous action with a learned 32-dimensional embedding of the same.
We bin relative distance into bins increasing in size exponentially with distance and relative angle into 5 degree bins before passing through embedding layer.

%\subsection{Rewards and Optimization}
%\label{sec:rewards}
The network is trained end-to-end using proximal policy optimization (PPO)~\cite{ppo}, following~\cite{habitat19iccv}.
All policies trained with this receive a reward $R_t$ at time step $t$ proportional to the distance they have moved towards the goal.
% \begin{equation*}
%         R_t=\begin{cases}
% s + \Delta_{t - 1} - \Delta_{t} + \lambda & \text{if goal is reached}
% \\ \Delta_{t - 1} - \Delta_{t} + \lambda & \text{otherwise}
% \end{cases},
% \end{equation*}
% where $s$ is the reward for a success, $\Delta_t$ is the agent's distance from the goal at time step $t$, and $\lambda$ a small time penalty.
We also include an auxiliary coverage reward equal to the change in the percentage of the ground truth scene graph observed by agent.
This helps to encourage the agent to explore and seek better views to complete more of scene graph.
% We train policies also include a small penalty for collisions with the environment, equal to $0.1\lambda$.
For more training details (\eg, learning rates, model architectures, \etc), we refer to \cite{anm20iclr}, \cite{habitat19iccv}.

\vspace{-0.1cm}
\paragraph{Evaluated Models. }
To evaluate our network, we modify and compare against two best-performing, end-to-end RL methods~\cite{anm20iclr}, i.e., \textbf{(1) RL + \approach: } An RL policy that replaces the ResNet18~\cite{Resnet} of ``RL + Res18''~\cite{anm20iclr} with our \approach policy architecture, followed by a GRU~\cite{cho14:gru}. \textbf{(2) RL + \approach + ProjDepth: } A baseline adapted from ``RL + Res18 + ProjDepth''~\cite{anm20iclr,exp4nav} that---in addition to replacing ResNet18 visual encoder with \approach---also projects the depth to an egocentric top-down map as input to the RL policy. Following~\cite{anm20iclr}, we pass the depth image through a three-layer CNN~\cite{habitat19iccv} before concatenating with output of \approach and passing to the GRU policy.

For fair comparison (as explained below), we also implement an RGB-D only baseline that uses ResNet50 identical to backbone of \approach (and used in works such as~\cite{ddppo}).
%Note we do not compare directly against Active Neural SLAM (ANS)~\cite{anm20iclr} due to the unavailability of its code during our implementation (ANS code was recently released). However, 
We believe our results demonstrate \approach can act as a modular component in any learning-based visual navigation system. We hope to demonstrate the compatibility of ours  with other approaches (e.g., Active Neural SLAM~\cite{anm20iclr}) in future.

\begin{figure}[]
    \centering
    %\vspace{-0.1cm}
    \includegraphics[width=0.37\textwidth, height=5.5cm]{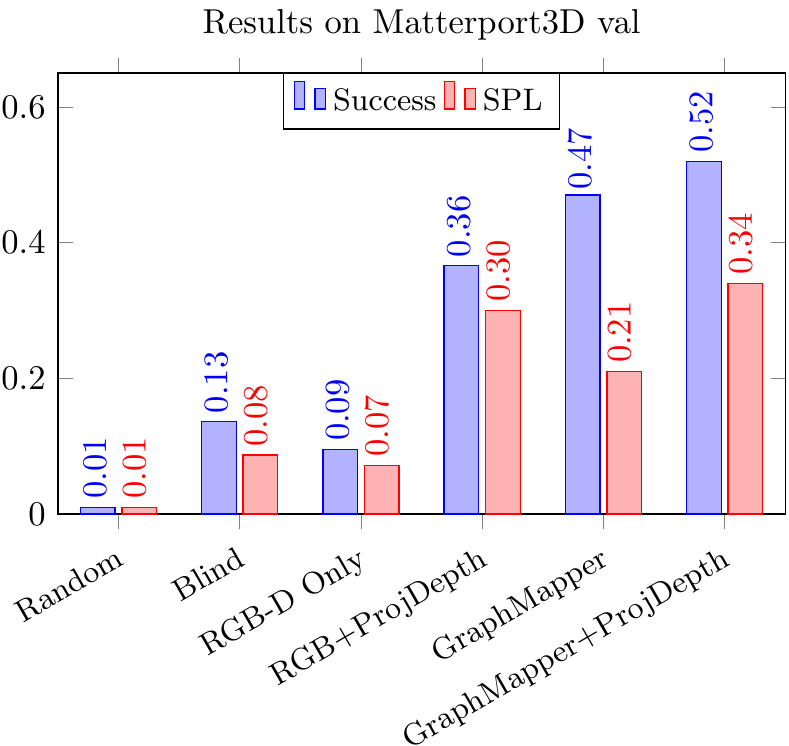}
    %\vspace{0.1cm}
    \caption{A comparison of several RL-based PointGoal solutions, with and without our proposed \approach scene graph encoder. 
    %All models are trained on Matterport3D for 14 million steps (frames of experience)
    %, and results are reported on 495 episodes from the Matterport3D val split.
    }
    \vspace{-0.3cm}
    \label{fig:quant}
\end{figure}

\vspace{-0.1cm}
\paragraph{Quantitative Results. }
In \Cref{fig:quant}, we present the results of several Reinforcement Learning (RL) models.
In addition to the Random (unlearned) and Blind (no visual observation) baselines, we show two visual navigation policies without (RGB-D Only and RGB+ProjDepth) and with (\approach and \approach+ProjDepth) our Scene Graph Transformer Network in the visual encoder.
These results were calculated on the Matterport3D validation split~\cite{habitat19iccv}, a set of $495$ navigation episodes %(starting location and target location pairs) in scenes 
not observed during training.
All agents were trained with an equal amount of experience---$14$ million steps on the Matterport3D train split---which has been shown to be sufficient experience for (depth-only) learning-based models to outperform traditional SLAM~\cite{habitat19iccv}.
From \Cref{fig:quant}, with this amount of training the ResNet50 model underperforms compared to the Blind agent.
While other works have shown that such architectures are able to perform quite well after significantly more training~\cite{ddppo} (99\% \sr after training for 2.5 billion observations), we observe even after $14$ million observations of the environment, the policy has not quite learned to effectively generate feature representations useful for navigation.
We show that addition of \approach leads to more efficient training, by comparing agents trained with an equal amount of experience, also leading to significant performance improvements on the given training budget. 

\begin{figure}[]
    \centering
    \includegraphics[width=0.42\textwidth, height=4.6cm]{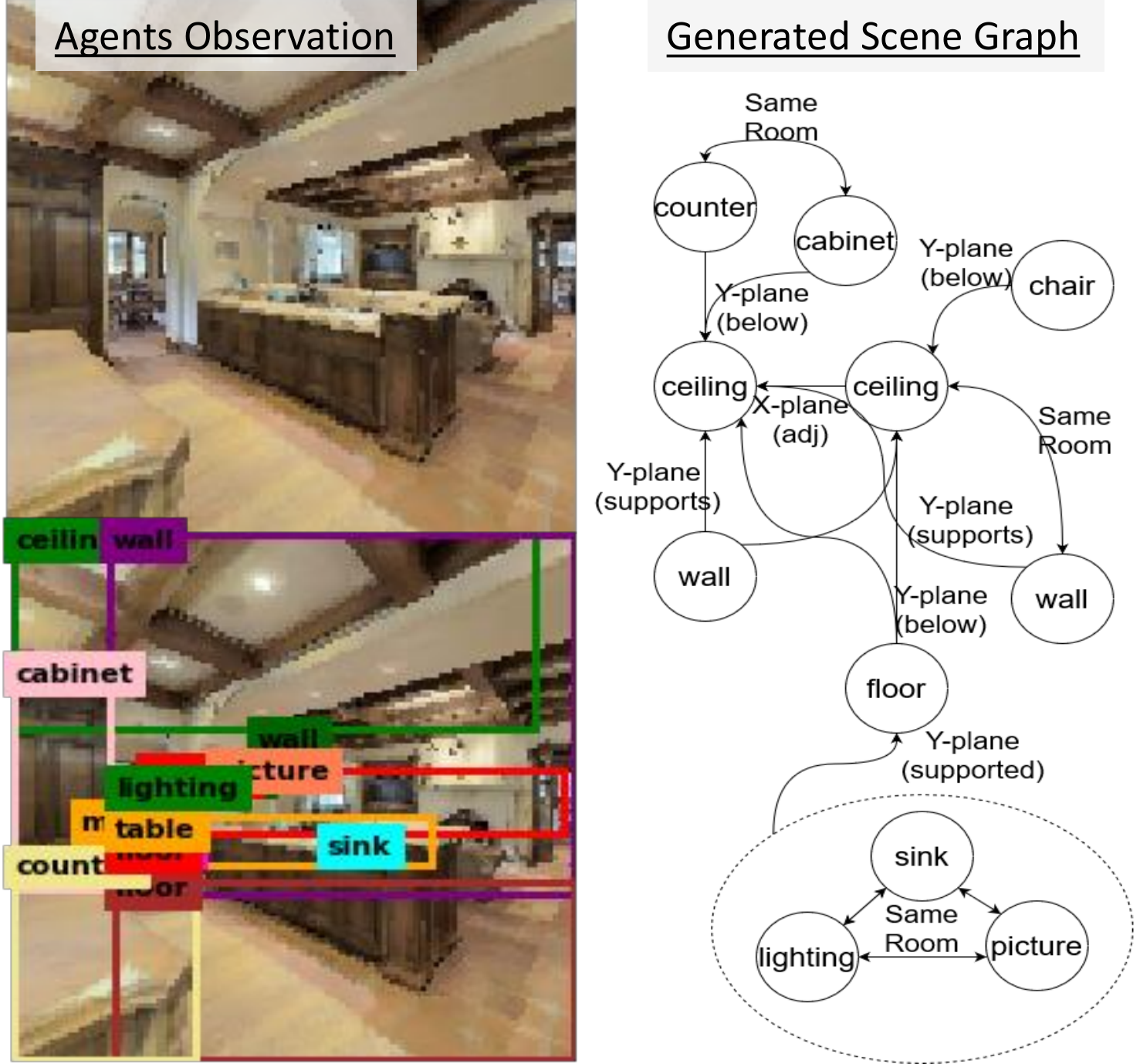}
    \vspace{0.1cm}
    \caption{Example of a scene graph generated by our method. We show the agent's observation (left top), the predicted node labels with their original bounding boxes (left bottom), and the generated graph (right). 
    %As there can be a large number of output edges, we visualize only the most-highly attended edges using the scores from the Scene Graph Transformer Network, for clarity.
    }
    \vspace{-0.3cm}
    \label{fig:sgg_example}
\end{figure}

On the other hand, RGB+ProjDepth, which supplements its visual observation with an accumulated, top-down egocentric map of the environment, is able to overcome these deficiencies and perform quite well.
%, particularly in terms of \spl.
Replacing the standalone ResNet50 encoder with \approach yields even greater improvements in terms of success, although its path efficiency, as measured by \spl, decreases.
It may be that the graph-based exploration reward alone overbiases the model towards taking longer paths, focusing on observing more of the graph structure. 
We note, however, that although the backbone network of \approach encodes similar visual priors (from pretraining on ImageNet and then MSCOCO object detection) as the RGB-D Only method, our Scene Graph Transformer Network is better able to refine the visual features with semantic-structural priors to increase the sample efficiency of the network, nearly doubling the \spl with the same amount of experience.
Furthermore, \approach+ProjDepth yields the best results, as structurally-refined visual features and egocentric map features are more closely aligned in encoding the agent's local 3D space, such that the policy can take effective advantage of each.

\vspace{-0.1cm}
\paragraph{Qualitative Results. }
In addition to effectively learning visual navigation policies, \approach also enables the generation of scene graphs of the environment. 
In \Cref{fig:sgg_example}, we present an example of a scene observed by the agent from Matterport3D and the scene graph generated internally by \approach.
\iffalse
\begin{figure}[tbh]
    \centering
    \includegraphics[width=0.8\columnwidth]{figures/graph_generation_example.png}
    \vspace{0.1cm}
    \caption{An example of a scene graph generated by our method. We show the agent's observation (left), the predicted node labels with their original bounding boxes (center), and the generated graph (right). 
    As there can be a large number of output edges, we visualize only the most-highly attended edges using the scores from the Scene Graph Transformer Network, for clarity.
    }
    \label{fig:sgg_example}
\end{figure}
\fi
For each node classified with high confidence ($p > 0.5$), we draw bounding box provided to the network and the label predicted.
On the right of \Cref{fig:sgg_example}, we draw the same nodes as part of the heterogeneous scene graph with a subset of the most confident evidence, chosen by high attention scores (top 10\%) from the meta-path selection in the Scene Graph Transformer Network. 
%As there can be a large number of output edges, we visualize only the most-highly attended edges using the scores from the Scene Graph Transformer Network, for clarity.
We label each edge with its predicted category (either ``same room'' or sharing one of the $x, y,$ or $z$ planes) as well as, where appropriate, a potential semantic explanation of said relationship.
For example, as in prior works~\cite{vgfm,scenegraphnet}, objects that are relatively co-planar both horizontally and vertically can be said to share a ``supporting--supported by'' relationship.
Furthermore, by accumulating these graphs over an entire navigation trajectory, \approach can be utilized for richer downstream tasks, such as mapping the environment into discrete semantic spaces (see supplementary).
%or providing high-level feedback about action selection (see \Cref{app_sec:interp}).

\subsection{Vision-Language Navigation } 
\label{sec:exp_il}

\iffalse
\begin{wraptable}{r}{0.46\textwidth}
%\scriptsize
\vspace{-0.4cm}
\centering
\caption{Experimental results to compare several Vision-Language Navigation agents on VLN-CE validation-seen set.}\label{tab:vln}
\vspace{0.15cm}
\resizebox{0.46\textwidth}{!}{%
\begin{tabular}{clccccccc}
\toprule
& & \multicolumn{5}{c}{\textbf{Evaluation Metrics}} \\ 
\cmidrule(){3-7}
Learning & Model & \textbf{SR}\uparrow & \textbf{SPL}\uparrow & \textbf{NDTW}\uparrow & \textbf{OS}\uparrow & \textbf{NE}\downarrow \\
\midrule
\multirow{2}{}{No} &  Random & 0.02 & 0.02 & 0.28 & 0.04 & 10.20 \\
&  Hand-Crafted  & 0.04 & 0.04 & 0.33 & 0.05 & 8.78  \\
  \midrule
\multirow{5}{}{Yes} &  RGB & 0.03 & 0.03 & 0.29 & 0.10 & 10.76 \\
&  Depth & 0.24 & 0.23 & 0.46 & 0.31 & 8.55 \\
& RGB-Depth & 0.25 & 0.24 & 0.45 & 0.35 & 8.54 \\
& \approach & 0.18 & 0.17 & 0.40 & 0.23 & 8.82 \\
& \approach-Depth & 0.27 & 0.26 & 0.47 & 0.35 & 8.37 \\
\bottomrule
\end{tabular}}
\end{wraptable} 
\fi

VLN task requires the agent to navigate in a 3D environment to a target location following language instructions. Our base network for performing this task is derived from sequence-to-sequence baseline described in~\cite{krantz2020navgraph}. Similar to Sec.~\ref{sec:exp_drl}, we replace visual encoder with our Scene Graph Transformer. The network is trained end-to-end using teacher-forcing IL training~\cite{williams1989learning, krantz2020navgraph}. Teacher-forcing minimizes the maximum-likelihood loss using ground-truth trajectories. We again include an auxiliary reward equal to the change in the percentage of the ground truth scene graph observed by the agent.
%We use VLN-CE dataset~\cite{krantz2020navgraph}

\renewcommand{\arraystretch}{0.94}
\begin{table}[]
%\scriptsize
\vspace{0.1cm}
\centering
\caption{Several VLN agents on VLN-CE validation-seen set.}\label{tab:vln}
%\vspace{0.15cm}
\resizebox{0.47\textwidth}{!}{%
\begin{tabular}{clcccc}
\toprule
& & \multicolumn{4}{c}{\textbf{Evaluation Metrics}} \\ 
\cmidrule(){3-6}
\textbf{Learning} & \textbf{Model} & \textbf{SR}$\uparrow$ & \textbf{SPL}$\uparrow$ & \textbf{NDTW}$\uparrow$ & \textbf{NE}$\downarrow$ \\
\midrule
\multirow{2}{*}{No} &  Random~\cite{krantz2020navgraph} & 0.02 & 0.02 & 0.28 & 10.20 \\
&  Hand-Crafted~\cite{krantz2020navgraph}  & 0.04 & 0.04 & 0.33 & 9.56  \\
  \midrule
\multirow{5}{*}{Yes} &  RGB~\cite{krantz2020navgraph} & 0.03 & 0.03 & 0.29 & 10.76 \\
&  Depth~\cite{krantz2020navgraph} & 0.24 & 0.23 & 0.46 & 8.55 \\
& RGB-Depth~\cite{krantz2020navgraph} & 0.25 & 0.24 & 0.45 & 8.54 \\
& \approach & 0.18 & 0.17 & 0.40 & 8.82 \\
& \approach-Depth & \textbf{0.27} & \textbf{0.26} & \textbf{0.47} & \textbf{8.37} \\
\bottomrule
\end{tabular}}
\vspace{-0.25cm}
\end{table} 

\paragraph{Quantitative Results. } In Table~\ref{tab:vln}, we compare several VLN baselines on VLN-CE dataset validation seen split~\cite{krantz2020navgraph}. The table is divided into No-Learning and Learning parts to aid our study. The compared models are: \textbf{1) No-Learning (i.e., Random and Hand-Crafted):} Agents do not process any sensor input and have no learned component \cite{krantz2020navgraph}. \textbf{2) Learning (i.e., Sequence to Sequence):} We use Sequence to Sequence (Seq2Seq) VLN baseline following~\cite{krantz2020navgraph, mattersim}. Seq2Seq agent employs a recurrent policy that takes a representation of instructions and visual observations (RGB, Depth, \approach) at each time step and then predicts an action. 

The agents are trained following~\cite{krantz2020navgraph} for at most $30$ epochs and the best performing models are selected. We use the Adam optimizer and a learning rate of $2.5\mathrm{e}{-4}$. 
We use ResNet50 model~\cite{Resnet} pretrained on ImageNet to extract RGB Image features and ResNet50 model pretrained on PointGoal task with DDPPO~\cite{ddppo} to extract Depth features. The models are implemented in PyTorch~\cite{paszke2019pytorch}, on top of Habitat-API~\cite{habitat19iccv} version $0.1.5$ and trained utilizing four RTX $2080$ Ti GPUs. 
%Please see~\cite{krantz2020navgraph} for more details.

We provide Seq2Seq models with different inputs in Table~\ref{tab:vln}. We observe that the RGB model only performs on par with the no-learning baselines. The Seq2Seq Depth baseline shows a large improvement due to having access to ground-truth depth information. We note that, although such explicit dense depth cues help the agent significantly to traverse effectively, the performance of the agent is unlikely to transfer well to real-world settings due to the noisy and sparse nature of sensor data available from existing depth-sensing technologies. On the other hand, \approach only depends on implicit depth (3D positions) and, hence, is more likely to transfer well. The \approach agent performs significantly better than the RGB agent (\eg, $0.18$ \vs $0.03$ in \sr) and \approach-Depth agent outperforms the RGB-Depth agent (\eg, $0.27$ \vs $0.25$ in \sr). Furthermore, \approach-Depth yields the best results, which again shows that \approach can act as a modular component to improve the performance of other models.

\begin{figure}[]
    \centering
    %\vspace{-0.1cm}
    \includegraphics[width=0.48\textwidth,, height=6.3cm]{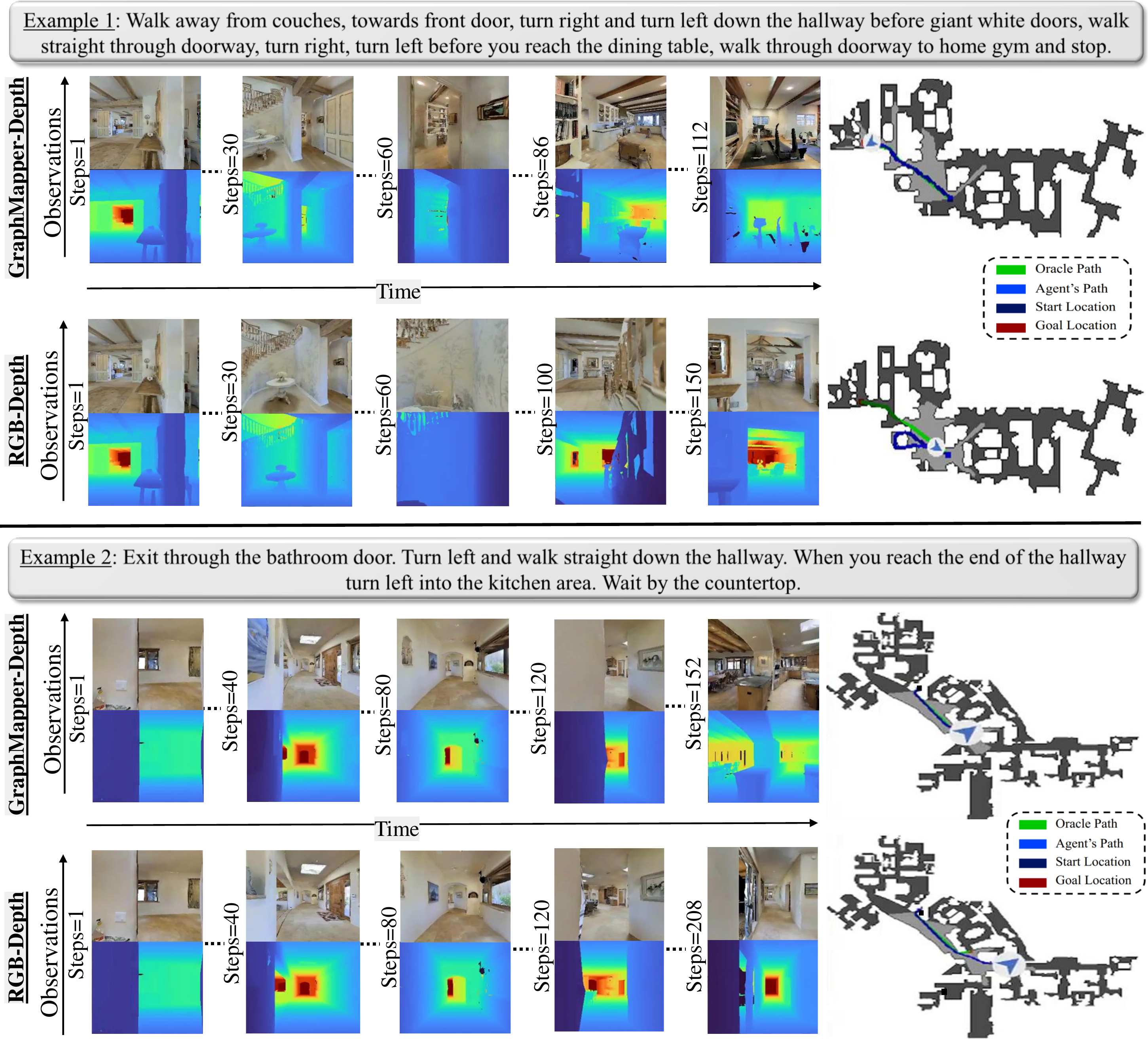}
    %\vspace{0.1cm}
    \caption{Figure shows instruction-following trajectories of our GraphMapper-Depth agent compared to RGB-Depth in environments within VLN-CE. Sample observations (i.e., RGB, Depth) seen by the agent are shown at each timestep. The top-down map (shown on the right) is not available to the agent and is only used for evaluation.
    }
    \vspace{-0.3cm}
    \label{qualitative_vln}
\end{figure}

\vspace{-0.1cm}
\paragraph{Qualitative Results. }
In \Cref{qualitative_vln}, we qualitatively compare our GraphMapper-Depth agent to RGB-Depth agent.
%report the performance of our GraphMapper-Depth agent compared to RGB-Depth agent with two examples. 
Both the examples represent success cases for our agent whereas RGB-Depth fails due to a lack of semantic understanding. In Example 1, we see RGB-Depth agent is unable to recognize the hallway before giant white doors, gets stuck near the staircase, and returns close to the start location. On the contrary, our agent successfully follows the complex sequence of instructions to finally perform STOP action after reaching its goal home gym. In Example 2, we see both the agents follow the instructions to reach close to the goal. The RGB-Depth agent reaches the kitchen but fails to identify the countertop and continues to take actions for several steps and incorrectly performs a STOP action in a hallway. Our GraphMapper-Depth agent is able to immediately understand that it is close to its goal location countertop after entering kitchen and performs a STOP action.

\section{Conclusion}
In this work, we have introduced a novel method for improving visual feature representations in learning-based visual navigation systems, \approach.
We have shown that using structurally and semantically refined features produced from our Scene Graph Transformer Network increases the navigational success and sample efficiency of these models.
Furthermore, we demonstrate that \approach is capable of generating scene graph representations of its environment that can be useful for explaining black box learning policies or performing other downstream mapping tasks.
In future work, we hope to explore the use of these scene representations for higher-level tasks, such as active SLAM or object finding, or for enabling better human interaction with autonomous systems.

{\small
\bibliographystyle{IEEEtran.bst}
\bibliography{egbib}
}
\end{document}

\maketitle

%{
% \renewcommand{\thefootnote}%
%    {\fnsymbol{footnote}}
% \footnotetext[2]{\scriptsize Center for Vision Tech., SRI International; {\tt\scriptsize firstname.lastname@sri.com}}
%}

\section{Experimental Details}
\label{app_sec:exp}

\vspace{0.4cm}
\subsection{PointGoal Navigation}

\vspace{0.2cm}
\subsubsection{Simulator and Dataset}
We use the Habitat simulator \cite{habitat19iccv} to conduct our experiments. 
This simulator supports different datasets such as Matterport3D~\cite{Matterport3D}, Gibson~\cite{Gibsonenv}, Replica~\cite{replica}, \etc. 
We focus our experiments on the Matterport3D dataset~\cite{Matterport3D}, which consists of 3D reconstructions of $90$ houses with a total of $2056$ rooms, as it exhibits the greatest episodic complexity and greatest density of scene objects, compared to the others.
Semantic segmentation masks are available for all the scenes in this dataset.
We make use of the semantic information to supervise node classification in our Scene Graph Transformer Network.
The autonomous agent is equipped with \rgb and \depth sensors, has a diameter \SI{0.2}{\meter} and a height of \SI{1.5}{\meter}, and can take four actions: \leftaction, \rightaction, \forwardaction, and \stopaction.
We make use of both depth sensor and actuation noise models based on real-world data that are provided as part Habitat~\cite{Choi_2015_CVPR,pyrobot2019}.

We use the train/val splits for Matterport3D as provided by~\cite{habitat19iccv}.
The set of scenes in each split is disjoint, so the agent is evaluated and tested on environments that were never seen during training.

\vspace{0.15cm}
\subsubsection{Evaluation Metrics}
\label{app_sec:eval_point}

For PointGoal navigation, we consider an episode a success if 
\begin{enumerate*}[label=\roman*)]
    \item the agent has taken fewer than $500$ steps,
    \item its distance to the goal is less than \SI{0.2}{\meter},
    \item and it has taken the \stopaction to indicate it has not randomly stumbled upon the target.
\end{enumerate*}
We report both the agent's success rate averaged over the test episodes, and we also report success weighted by path length (\spl) metric, as has become standard practice~\cite{Anderson2018OnEO}. 
% \spl is a measure of the efficacy of the policy's path taken to the goal compared against the ground truth shortest path.
% Let the number of evaluation episodes be $N$, the length of the shortest path $s$, and the actual distance covered by the agent $d$. Then, \spl is calculated as follows,
% \begin{equation*}
% \spl = \frac{1}{N} \Sum{\mathbbm{1} \{ i  \text{ is successful}\}  \frac{s_{i}}{\max \left(d_{i}, s_{i}\right)}}{i,1,N}.
% \label{eq:SPL}
% \end{equation*}

\vspace{0.3cm}
\subsection{Vision-Language Navigation}

\vspace{0.1cm}
\subsubsection{Simulator and Dataset}
We use VLN-CE dataset provided by Krantz et al.~\cite{krantz2020navgraph} and Habitat simulator~\cite{habitat19iccv} to perform our experiments. VLN-CE dataset is constructed by transferring trajectories and instructions from nav-graph-based Room-to-Room dataset (which is based on Matterport3D~\cite{Matterport3D}) to the continuous
setting in VLN-CE in Habitat simulator. The Matterport3D dataset ~\cite{Matterport3D} is a collection of $90$ different indoor environments captured through 10.8k RGB-D panoramic views. The Room-to-Room dataset~\cite{mattersim} contains 7189 trajectories with three human instructions. 

Following an agent inside Matterport3D environment, the VLN-CE dataset provides $4475$ navigable trajectories (available in Habitat Simulator). Each trajectory is associated with 3 instructions annotated by humans.  The VLN-CE trajectories includes on average $55$ actions along path, making the problem richer and realistic compared to prior VLN datasets (e.g., Room-to-Room) which includes on average trajectory length of $4-6$.

The VLN-CE dataset provides the following low-level actions for each instruction-trajectory pair for navigation inside Habitat Simulator: \texttt{move forward} ($0.25m$), \texttt{turn-left} or \texttt{turn-right} ($15 deg.$) and \texttt{stop}. We used the train, and validation splits following~\cite{krantz2020navgraph} in experiments.

\vspace{0.2cm}
\subsubsection{Evaluation Metrics}
We use the following standard visual navigation metrics to evaluate our agents as described in previous works~\cite{Anderson2018OnEO, 49206, mattersim}: Success rate ($\textbf{SR}$), Success weighted by path length ($\textbf{SPL}$), Normalized Dynamic Time Warping ($\textbf{NDTW}$), and Navigation Error ($\textbf{NE}$). A navigation episode is considered a success if the agent has taken a \stopaction signal stops $3m$ of the goal. Similar to Sec.~\ref{app_sec:eval_point}, We report the agent's success rate ($\textbf{SR}$) averaged over the test episodes and success weighted by path length (\spl) metric. We report $\textbf{NDTW}$ which is used as an estimate of how agent's path deviate from the reference trajectory. We also report navigation error ($\textbf{NE}$) which calculates average navigation error from goal in meters. Following prior works \cite{Wang2019ReinforcedCM, krantz2020navgraph, Anderson2018OnEO}, we use \spl and \sr as the primary metrics to evaluate the navigation performance of our agent. 

\iffalse
\subsubsection{Quantitative Results}
We have reported quantitative results on VLN-CE dataset validation seen split on Table 1 (Sec. 5.2) of the main paper. Here, we report results on VLN-CE dataset validation unseen split on Supp. Table~\ref{tab:vln2}. We observe similar trends to what observed in the validation seen results.

\begin{table}[h]
\scriptsize
%\small
\centering
\caption{A comparison of several VLN agents on VLN-CE validation-unseen set.}\label{tab:vln2}
\vspace{0.25cm}
\begin{tabular}{clcccc}
\toprule
& & \multicolumn{4}{c}{\textbf{Evaluation Metrics}} \\ 
\cmidrule(){3-6}
\textbf{Learning} & \textbf{Model} & \textbf{SR}$\uparrow$ & \textbf{SPL}$\uparrow$ & \textbf{NDTW}$\uparrow$ & \textbf{NE}$\downarrow$ \\
\midrule
\multirow{2}{*}{No} &  Random~\cite{krantz2020navgraph} & 0.03 & 0.02 & 0.30 & 9.51 \\
&  Hand-Crafted~\cite{krantz2020navgraph}  & 0.03 & 0.02 & 0.30 & 10.34  \\
  \midrule
\multirow{5}{*}{Yes} &  RGB~\cite{krantz2020navgraph} & 0.04 & 0.04 & 0.31 & 9.89 \\
&  Depth~\cite{krantz2020navgraph} & 0.17 & 0.15 & 0.41 & 9.09 \\
& RGB-Depth~\cite{krantz2020navgraph} & 0.20 & 0.18 & 0.43 & 8.94 \\
& \approach & 0.11 & 0.10 & 0.36 & 9.41 \\
& \approach-Depth & \textbf{0.21} & \textbf{0.19} & \textbf{0.43} & \textbf{8.88} \\
\bottomrule
\end{tabular}
\vspace{-0.15cm}
\end{table} 
\fi

%\vspace{5cm}

%\vspace{5cm}

\vspace{0.8cm}
\section{Applications}

\vspace{0.3cm}
\subsection{Unsupervised Room Separation}
\label{app_sec:rooms}
\begin{figure}[]
%\vspace{0.1cm}
\begin{center}
\includegraphics[width=0.34\textwidth]{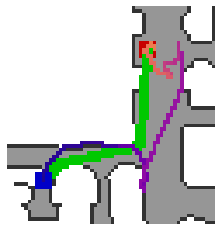}
\vspace{0.15cm}
\includegraphics[width=0.48\textwidth]{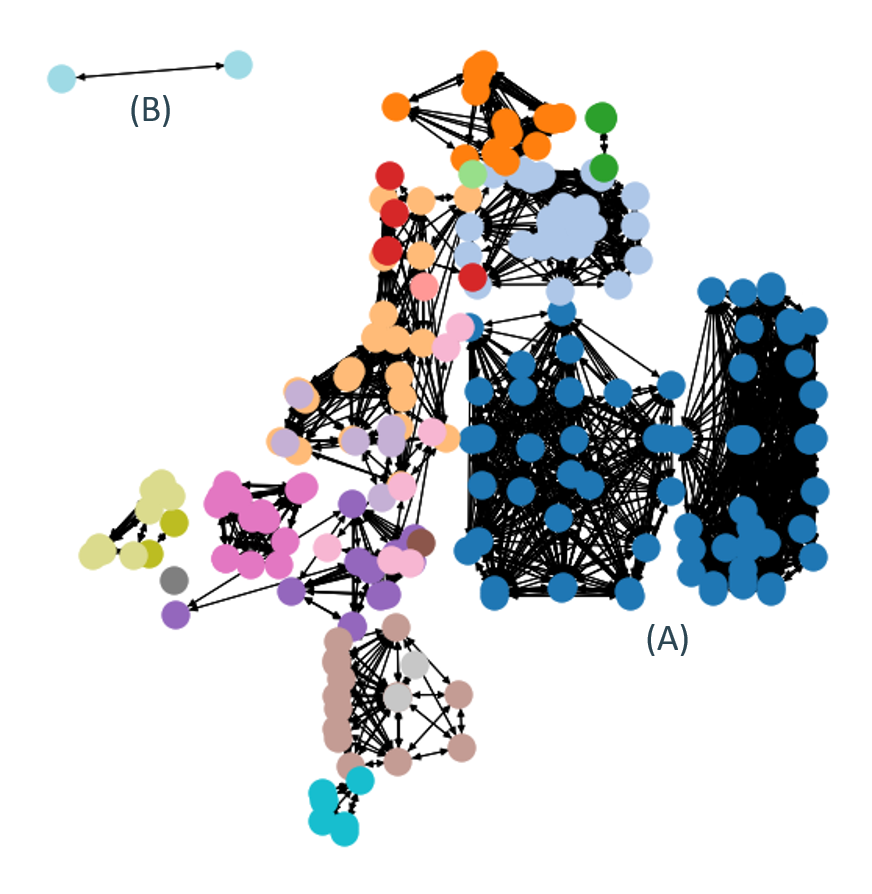}
\end{center}
%\vspace{0.2cm}
\caption{An example of a scene graph accumulated by the agent (bottom) as it moves follows a path through the environment (top). The node labels and the predicted ``same room'' edges are depicted. The graph nodes are arranged by their ground truth coordinates, and each connected component is colored differently to demonstrate how the agent automatically discretizes the space into regions.}
\label{fig:sg_rooms}
\end{figure}

After object classification and semantic scene segmentation, one of the most useful mid-level visual features for an agent to perform visual navigation is scene classification~\cite{MidLevelVisualRep}; that is, not only \emph{where} is the agent but in \emph{what kind} of room is it?
As agent policies move towards higher-level autonomy, the answer to this type of question will greatly influence its decision-making, such as what kinds of object it is able to retrieve or the potential actions it can undertake in the given space.
This sort of higher-order understanding is also important for tasks which require human interaction with the agent, such as vision-and-language navigation (VLN)~\cite{EmbodiedVLN}, where an agent must navigate by following human directions.
Humans tend to speak in high order, semantic commands (``go to the kitchen and bring me an apple'') rather than low-level actions (``move forward \SI{3}{\meter}, rotate arm \SI{30}{\degree}''); thus, an agent must accumulate this high-level understanding of the environment through its experience.

\vspace{0.2cm}
Our \approach enables one possible instance of this ability.
As an agent traverses from its starting location to the goal, we store the scene graph predicted from each from and then fuse like nodes by their approximate world coordinates.
We then keep the homogeneous graph consisting of only the nodes and the predicted ``same room'' edge.
In this way, each connected  component of the graph should correspond to a unique room in the house environment.
In \Cref{fig:sg_rooms}, we depict the resulting graph, assigning each connected component a unique color.
As can be seen, the clusters seem mostly reasonable, with nearby nodes sharing a region and only a few minor errors at the edges.
Particularly, the two rooms at the right of the figure (marked ``(A)'') have been merged into one component, with a node labeled ``wall'' seemingly connecting the two. 
This is an artifact of the way Matterport3D annotates the ``wall'' class---with all four walls of a room being a single instance---often leading to irregularities in our ground truthing.
On the other hand, at the top of the figure (marked ``(B)'') we see the agent correctly handling an inconsistency in the underlying mesh.
Namely, from its starting position, it has a correctly identified a door and corresponding wall---observed through a window---that are not part of the navigable space of the mesh.
However, given their geometric relationship with the scene and the rest of the entity observed, the agent correctly segregates these nodes to a distant, separate graph component.

\vspace{0.5cm}
\subsection{Path/Action Interpretability}
\label{app_sec:interp}
\begin{figure*}[]
\centering
\includegraphics[width=0.9\textwidth]{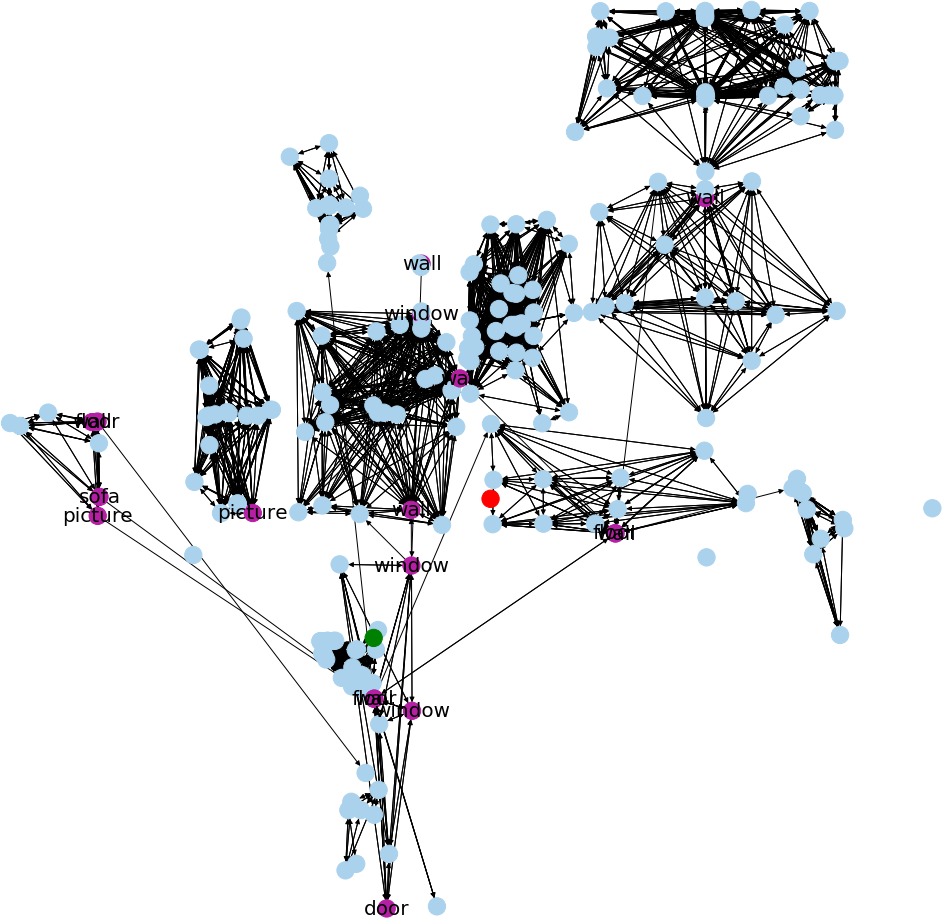}
\caption{The scene graph accumulated by the agent, with the most highly attended nodes at each step highlighted. The agent's starting position and target location are highlighted in green and red, respectively. At each time step, we find the outgoing node with highest attention from the Scene Graph Transformer Network and highlight it in purple, along with its label.}
\label{fig:sg_showing_path}
\end{figure*}

Finally, we demonstrate one additional use of this method with great potential for future applications.
In \Cref{fig:sg_showing_path}, we show the same graph as in \Cref{fig:sg_rooms}, with the ``same plane'' edges added.
We also add additional nodes added to mark the starting location (green) and ending location (red) of the agent.
As the agent navigates to its target location, at each time step, we find the outgoing node in the current scene graph with the highest attention score from the Scene Graph Transformer Network and highlight it in purple.
This affords an indication of which parts of the scene \approach most focuses on to generate the policy features and make its decisions.
Thus, in a way, the nodes attended to in such a manner can act as an indication of which parts of the scene most contributed to the agent's decision at each timestep.
In most cases, we see these are structural regions of the scene that either provide potential egress points---``door'' and (to a lesser extent) ``window''---or that shape the space and act as potential obstacles---such as ``wall,'' ``picture'' (likely by way of ``wall''), and ``sofa.''
Such an example of scene understanding offers many potential applications in visual navigation, particularly as the agent moves towards more complicated tasks.
For example, selection and memory of previously-observed egress points could help to automate the selection of recovery waypoints or alternate routes, should the agent become stuck in a dead end with from its previous decisions.
Alternatively, these attention scores could be used to help an agent selected new long-term goals to improve automatic exploration~\cite{anm20iclr} or to increase its ability to locate specific parts of the scene, as in ObjectGoal or vision-and-language navigation~\cite{EmbodiedVLN}.

\vspace{2cm}

{\small
\bibliographystyle{IEEEtran.bst}
\bibliography{egbib}
}